\documentclass{article}
\usepackage{spconf,amsmath,graphicx}

\usepackage{booktabs} 
\usepackage{hyperref}
\usepackage{algorithm}
\usepackage{algpseudocode}
\usepackage{tabularx} 
\usepackage{multirow}
\usepackage{enumitem}
\usepackage{amssymb}
\usepackage{mathtools}
\usepackage{amsthm}
\usepackage{svg}
\usepackage{xcolor}
\usepackage{float}
\usepackage{adjustbox}

\title{MetaDIP: Accelerating Deep Image Prior with Meta-Learning}
\name{Kevin Zhang, Mingyang Xie, Maharshi Gor, Yi-Ting Chen, Yvonne Zhou, Christopher A. Metzler }
\address{University of Maryland, College Park}

\begin{document}
\ninept
\maketitle
\begin{abstract}
Deep image prior (DIP) is a recently proposed technique for solving imaging inverse problems by fitting the reconstructed images to the output of an untrained convolutional neural network. Unlike pretrained feedforward neural networks, the same DIP can generalize to arbitrary inverse problems, from denoising to phase retrieval, while offering competitive performance at each task. The central disadvantage of DIP is that, while feedforward neural networks can reconstruct an image in a single pass, DIP must gradually update its weights over hundreds to thousands of iterations, at a significant computational cost. In this work we use meta-learning to massively accelerate DIP-based reconstructions. By learning a proper initialization for the DIP weights, we demonstrate a 10$\times$ improvement in runtimes across a range of inverse imaging tasks. 
Moreover, we demonstrate that a network trained to quickly reconstruct faces also generalizes to reconstructing natural image patches.
\end{abstract}
\vspace{-10pt}
\section{Introduction}
\vspace{-5pt}
\label{sec:intro}

In recent years, deep learning has emerged as a powerful method for solving inverse problems in imaging. 
Many methods which train deep convolutional neural networks to take in measurements and output an estimate of the ground truth have been developed for problems in the fields of medical imaging, computational microscopy, computational photography, and more~\cite{ongie2020deep}.
Generally these methods must be trained on a huge dataset, which may not be accessible in the context of applications like scientific imaging. Moreover, these models are often problem specific and solving a new problem requires training an entirely new network.

Alternatively, Ulyanov et al.~propose to view the structure of deep convolutional neural networks as a generative prior over natural images~\cite{8579082}. 
This approach, referred to as Deep Image Prior (DIP), requires no training data and has been used successfully to solve a range of both linear~\cite{van2018compressed} and non-linear~\cite{liu2021solving} inverse problems. 
However, a key limitation of DIP is that it requires many iterations and excessive computations to solve a single inverse problem.

Inspired by work showing that meta-learning can speed up the speed at which deep neural networks solve regression tasks \cite{tancik2020meta}, we show that the speed at which DIP solves inverse problems in imaging can be improved using meta-learning. We first use meta-learning to find an optimal initialization for removing additive  white Gaussian noise from celebA faces. We then demonstrate that this same initialization improves performance across a variety of inverse problems and datasets. 
Our contributions can thus be summarized as follows:

\begin{itemize}
    \item Task agnostic meta-learning: We demonstrate that we can use meta-learning to optimize a network to denoise and then apply it to solve the compressive sensing and compressive phase retrieval inverse problems.
    \item Class agnostic meta-learning: We demonstrate that we can use meta-learning to optimize a network to reconstruct faces and then use this same network to reconstruct natural image patches.
    \item Competitive performance: We demonstrate performance and runtimes competitive with state-of-the-art plug and play methods.
    \end{itemize}
\vspace{-10pt}
\section{Related work}
\vspace{-5pt}

\subsection{Deep Image Prior}
Recently, there has been considerable interest in using untrained neural networks as generic image priors~\cite{8579082,heckel2018deep}.
These methods, exemplified by DIP, work by parameterizing an image as the output of a neural network and then optimizing the weights of this network so as to minimize the loss between predicted and observed measurement.
These methods offer strong performance across a variety of inverse problems in imaging such as superresolution, inpainting,  deblurring~\cite{8579082}, compressive sensing~\cite{van2018compressed}, phase retrieval~\cite{bostan2020deep}, computer generated holography~\cite{liu2021solving}, and more. In addition, by learning simple Gaussian priors on the distributions of DIP's weights, one can improve its performance~\cite{van2018compressed}. However, because recovering an image with DIP requires optimizing the network's weights over thousands of iterations, DIP is slow and computationally expensive. 

\subsection{Meta-Learning Neural Representations}

Meta-learning is defined as a set of methods which learn to learn, or optimize the learning process of a learning algorithm. 
In other words, the goal of meta-learning is to train a model in  a way such that it can adapt to new tasks and data quickly. 
The most well-known meta-learning algorithm is Model Agnostic Meta-Learning (MAML).
MAML learns a parameter initialization from which a model can adapt to a new task and data in a small number of gradient steps by differentiating through gradient steps to compute a meta-gradient and optimizing the initialization with this meta-gradient~\cite{finn2017model}. 


Tancik et al.~recently demonstrated that meta-learning can be used to find a set of weights that allow one to quickly fit coordinate-based neural representations of images and neural rendering fields~\cite{tancik2020meta}. 
MAML has also been applied to quickly adapt to neural representations of surfaces~\cite{Sitzmann2020MetaSDFMS} and lumigraphs~\cite{bergman2021metanlr}.  
While coordinate-based representations of images can be used to solve inverse problems in largely the same manner as DIP~\cite{sitzmann2019siren}, we will demonstrate in Section~\ref{sec:experiments} that these networks impose weak priors and are less effective than DIP at solving general inverse problems in imaging.

\subsection{Meta-learning for Inverse Problems in Imaging}
In~\cite{pmlr-v97-wu19d} Wu et al.~used meta-learning to update the weights of a generative network so that one could quickly solve a particular compressive sensing problem involving a fixed and learned measurement matrix. Our work goes significantly beyond this one by demonstrating that the same meta-learned initialization can generalize across a variety of different inverse problems (denoising, compressive sensing, and compressive phase retrieval) and data distributions (faces and natural image patches).

\vspace{-10pt}
\section{Proposed Method}
\vspace{-5pt}
\subsection{Solving inverse problems with neural representations}

Our goal is to recover the ground truth signal $x \in \mathbb{R}^n$ from a measurement $y = M(x, e) \in \mathbb{R}^m$, where $M$ represents a measurement process and $e \sim \mathcal{D}$ is noise coming from some distribution $\mathcal{D}$. 
This model is general. For instance, in the context of removing additive white Gaussian noise, the forward model is $M=I \in \mathbb{R}^{n \times n}$, where $I$ denotes an $n$-by-$n$ identity matrix, and $e \sim N(0, \sigma^2)$.

To apply neural representations to solve such an inverse problem, we solve the optimization problem 
\begin{equation} \label{eq:1}
    \theta^* = \min_{\theta} ||M(f_{\theta}(z))-y||_2^2,
\end{equation}
where $f$ is our choice of deep neural network architecture, $\theta$ are the model parameters, and $z \in \mathbb{R}^d$ is a fixed randomly initialized latent variable~\cite{8579082}. 
Our estimate of $x$, $x^* \in \mathbb{R}^n$ is then $x^* = f_{\theta^*}(z)$.
Optimizing this objective can be viewed as imposing a generative prior on $x^*$, with the idea that the structure of $f$ is suitably chosen so that it is biased towards producing natural images. 

\subsection{Meta-learning}
Optimizing Equation \eqref{eq:1} using conventional stochastic gradient descent-based methods is slow. 
To overcome this difficulty, we apply meta-learning. 
In the formulation of meta-learning relevant here, the goal is to find a set of initial weights $\theta_0^*$ such that the expected loss $L(\theta_m)$, in our case, \eqref{eq:1}, over a distribution of tasks $\mathcal{T}$ is minimized after $m$ steps of optimizing $L$ over sampled tasks $T \sim \mathcal{T}$, giving us the following optimization problem \cite{tancik2020meta}:
\begin{equation}\label{eq:2}
    \theta_0^* = \arg \min_{\theta_0} \mathbb{E}_{T \sim \mathcal{T}}[L(\theta_m(\theta_0, T))].
\end{equation}
One way of solving Equation \eqref{eq:2} is to apply Model Agnostic Meta-Learning (MAML). 
In MAML, the process of calculating the weights $\theta_m$ from the current initialization is referred to as the inner loop. 
To update this initialization, we wrap the inner loop in another layer of optimization, referred to as the outer loop.
More specifically, we apply gradient descent to optimize the loss associated with $\theta_m$ on task $T$ with respect to the initialization as 
\begin{equation}
    \theta_0^{j+1} = \theta_0^j - \beta \nabla_{\theta}L(\theta_m(\theta, T_j))\vert_{\theta=\theta_0^j}.
\end{equation}

\begin{figure}[t]
    \centering
    \includegraphics[width=\linewidth]{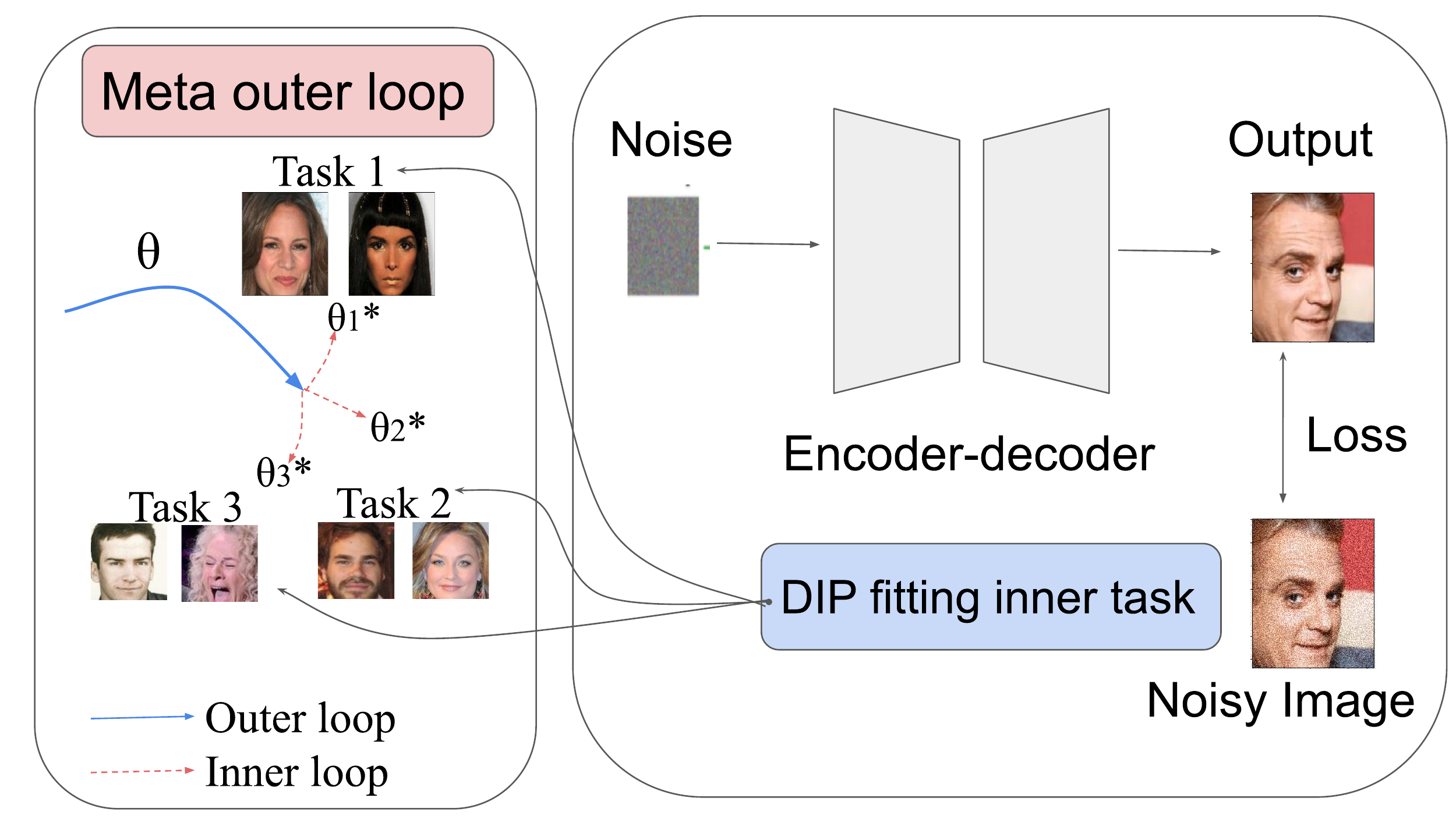}
    \caption{An illustration of the proposed method. By differentiating through the optimization process used to fit DIP, the inner task, in the outer loop we can use gradient descent to optimize an initialization from which DIP can fit natural images quickly. 
    }
    \label{fig:proposed}
\end{figure}

While it is possible to learn a different initialization for each inverse problem, we instead investigate if it is possible to learn an initialization for just one inverse problem, additive white Gaussian denoising, that is also useful for solving other unseen inverse problems. 
To adapt MAML to denoising, we sample the tasks as pairs of noisy and clean images, $y$ and $\tilde{y} = y+n$, where $n \sim N(0, \sigma^2)$ is additive white Gaussian noise and use an L2 reconstruction loss as our loss function. 
A visualization of our method is presented in Figure \ref{fig:proposed}.

\label{sec:method}

There is a risk that meta learning results in a deep image prior specific to a certain class of data, thereby giving up the data agnostic nature of DIP that made it so appealing in the first place. Fortunately, as we will demonstrate, the initializations found with meta-learning generalize and can effectively solve inverse problems involving unseen images categories. 

\subsection{Additional Details}
Because neural networks tend to be biased towards producing low frequencies, resulting in blurry outputs~\cite{pmlr-v97-rahaman19a}, we append a Fourier feature mapping to our input to enhance the ability of our networks to model high frequencies~\cite{NEURIPS2020_55053683}. 
We also improve the inner loop optimization process by learning a per-step learning rate~\cite{antoniou2018train}, adding a momentum  term to the gradient~\cite{Rumelhart:1986we}, and adding a fixed total variation penalty to the reconstruction loss term~\cite{8682856}. 
When fitting DIP without meta-learning in our baselines, we jitter the latent variable with additive white Gaussian noise and take the exponential moving average of the intermediate outputs over all the fitting iterations as our final reconstruction \cite{8579082}. 
\vspace{-10pt}




\section{Experiments}
\vspace{-5pt}
\label{sec:experiments}

We experimentally compare the performance of meta-initialized DIP (MetaDIP) with several baselines for denoising, compressive sensing, and compressive phase retrieval.

\begin{figure}[t]
    \centering
    \includegraphics[width=\linewidth]{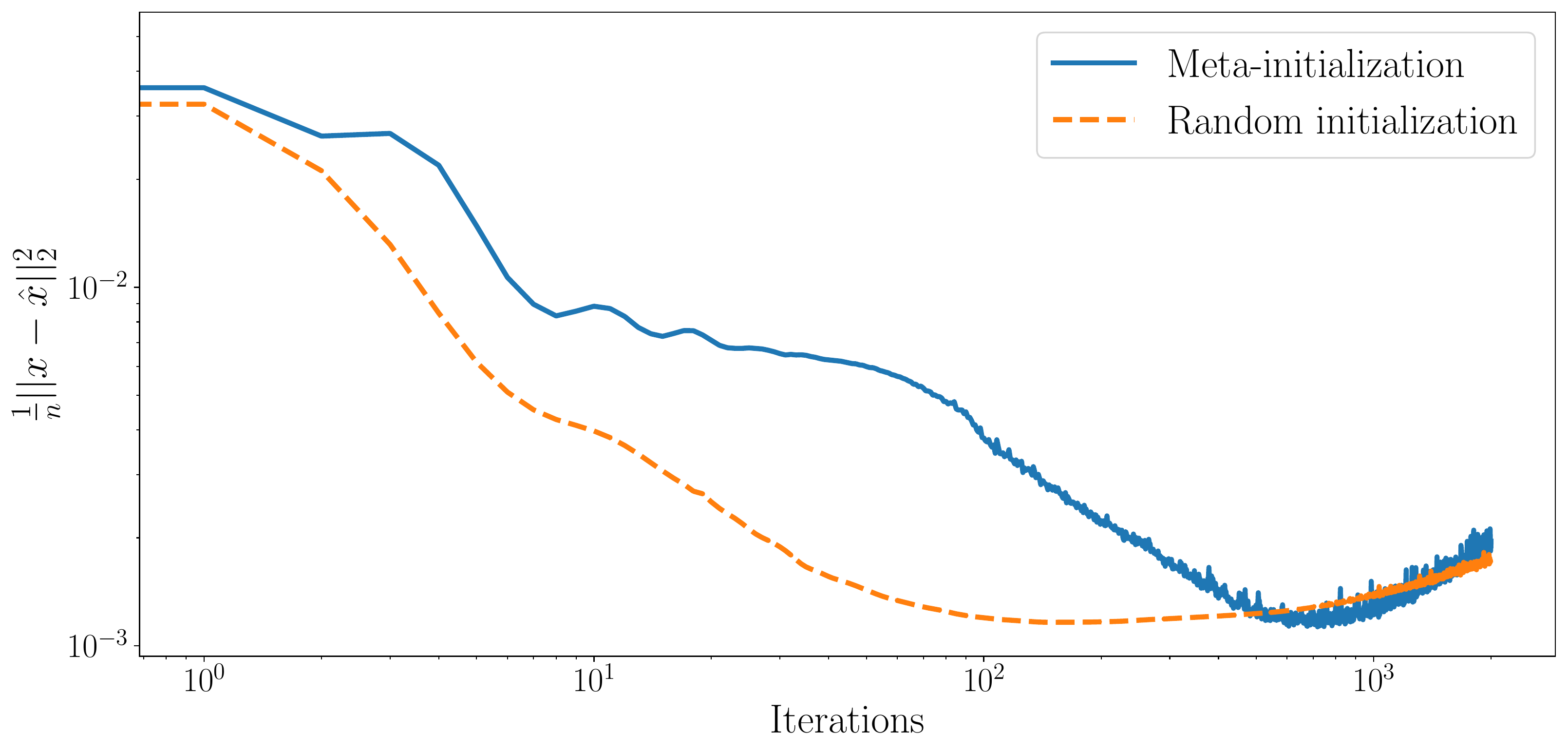}
    \vspace{-20pt}
    \caption{Convergence rates of DIP with a random initialization and a meta-learned initialization. DIP with the meta-learned initialization converges an order of magnitude faster.}
    \label{fig:Convergence}
\end{figure}

\subsection{Datasets}
\label{subsec:datasets}
\noindent \textbf{CelebFaces Attributes Dataset (CelebA)}~\cite{liu2015faceattributes} is a large-scale face attributes dataset with more than 200K celebrity images. For this task, we use $128 \times 128$ center-cropped images to train and test on.

\noindent \textbf{Natural image patches}: 
We curate a high-quality image dataset that combines multiple high resolution image datasets: the Berkeley Segmentation Dataset~\cite{MartinFTM01}, the Waterloo Exploration Database~\cite{ma2017waterloo}, the DIV2K dataset~\cite{Agustsson_2017_CVPR_Workshops}, and the Flick2K dataset~\cite{Lim_2017_CVPR_Workshops}~\cite{9454311} and sample random $128 \times 128$ patches from it to test on.


\subsection{Training/Architecture Details}
For the DIP experiments, we use a conventional UNet encoder-decoder architecture with skip connections~\cite{8579082}.
We learned meta-initializations with the CelebA dataset, where we sampled additive white Gaussian noise at $\sigma=25$. 
During meta-learning we performed 20 inner loop iterations (GPU memory limited the number of iterations). At inference time we performed 50 iterations. 
We train all of our meta initializations with the Adam optimizer \cite{DBLP:journals/corr/KingmaB14}. 


\subsection{Baselines}

\subsubsection{Descriptions}
For denoising, we compare MetaDIP against BM3D, a classical unsupervised denoising algorithm ~\cite{10.1117/12.643267}; DRUNet, a state-of-the-art deep CNN denoiser ~\cite{9454311}; Vanilla DIP, the original DIP \cite{8579082}; and MAML initialized SIREN, which we will call here MetaSIREN \cite{sitzmann2019siren}.

For the compressive inverse problems, we compare MetaDIP to Vanilla DIP, MetaSIREN, and Plug-and-Play Alternating Direction Method of Multipliers (PnP ADMM)~\cite{venkatakrishnan2013plug} with the BM3D and DRUNet denoisers. 
PnP ADMM is an iterative algorithm for solving imaging inverse problems. It works by iteratively minimizing a data fidelity term, which measures agreement of the current reconstruction with observed measurements; imposing a prior with an off-the-shelf denoising algorithm; and performing a dual update. 
PnP algorithms offer state-of-the-art performance on a range of inverse imaging tasks~\cite{9454311}, but they are notoriously difficult to tune~\cite{wei2020tuning}.


\subsubsection{Details}
For the SIREN experiments, we use a conventional feedforward neural network, but with sinusoidal activations~\cite{sitzmann2019siren}. Like MetaDIP, we meta-learn MetaSIREN with 20 inner loop iterations and used 50 inner loop iterations at test time. We train MetaSIREN with the same recipe as MetaDIP, with the exception that we do not learn the learning rate because we found that it made the training unstable and caused training to diverge.
Besides changing the number of iterations used, no other hyperparameters are changed at test time for MetaSIREN and MetaDIP.
Additionally, the hyperparameters for MetaSIREN and MetaDIP are the same across all inverse problems. In all the tasks, Vanilla DIP results are reported after $50$ iterations and $1000$ iterations and the intermediate outputs are combined using an exponential moving average to form the final output. 

For compressive sensing, the data fidelity term is minimized by using a least-squares update. 
For compressive phase retrieval, the data fidelity term is minimized by performing 10 subgradient descent steps with respect to the squared error loss . 
Notably, unlike our meta-learning approaches, the PnP ADMM-based methods require a time-consuming uniform grid search to properly set their hyperparameters, and they need different hyperparameters set for each compressive inverse problem.
Specifically, we searched over a 3-by-3-by-3 search grid to set the denoising strength, penalty parameter, and the number of iterations to use. 

\subsection{Denoising}

\begin{figure}[t]
    \centering
    \includegraphics[width=\linewidth]{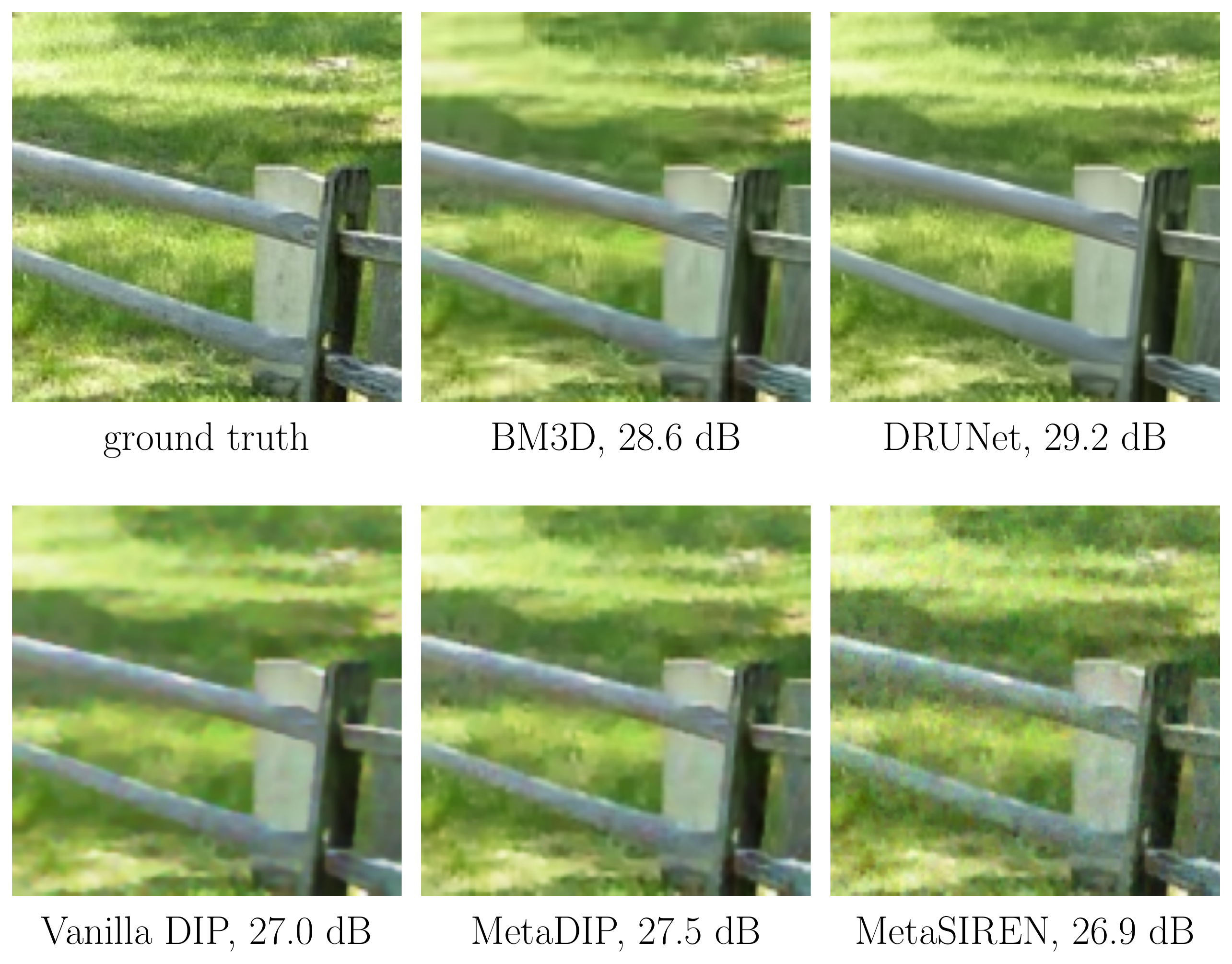}
    \vspace{-20pt}
    \caption{
    Various methods for denoising at $\sigma=25$, compared with respect to peak-signal-to-noise ratio (PSNR) on natural image patches. }
    \label{fig:patches}
\end{figure}

\setlength{\tabcolsep}{4pt}
\begin{table}[t]
	\ninept
	\parbox{\linewidth}{
	\centering
	\begin{tabular}{lrr|rr|r}
  \toprule
		\multirow{2}{*}{Method}& 
		\multicolumn{2}{c|}{$\sigma=15$} & \multicolumn{2}{c|}{$\sigma=25$} \\
		\cmidrule{2-6} 
		& \shortstack{PSNR \\ (faces)} &  \shortstack{PSNR \\ (patches)}  & \shortstack{PSNR \\ (faces)} & \shortstack{PSNR \\ (patches)} & Time  \\
		\midrule
		BM3D & 34.4 & 34.4 & 31.9  & 31.8 & 1.66 \\
		DRUNet &  35.5 & 34.8 & 33.2  & 32.5 & 0.01 \\ 
		DIP (1000) & 31.6 & 30.7 & 30.6  & 30.1 & 21.6 \\
		DIP (50) & 10.1 & 10.9 & 10.1  & 10.9 & 1.04 \\
		MetaSIREN &31.4  & 30.4 & 30.0 & 29.0 & 0.19 \\
		MetaDIP & 32.0 & 31.7 & 30.7 & 30.4 & 0.89 \\
		\bottomrule
	\end{tabular}%
	\caption{Average PSNRs (dB) and runtimes (seconds) of various methods for denoising $128 \times 128$ natural image patches. The times are averaged together as there was not a significant dependence of runtime on noise level for each method.}
	\label{tab:denoising_patch}%
	}
\hfill
\end{table}%

We first compare the convergence rates (with respect to the normalized mean squared error) of meta-initialized DIP and standard DIP in Figure~\ref{fig:Convergence}. Meta-initialized DIP converges an order of magnitude faster. Next we we present qualitative results for denoising  natural image patches in Figure \ref{fig:patches}. Quantitative results on denoising both faces and natural image patches are summarized in Table \ref{tab:denoising_patch}. 

Although MetaDIP does not outperform BM3D, it outperforms MetaSIREN and demonstrates qualitatively good denoising results.
Despite being explicitly trained on human face and not natural images patches, the learned initializations  are still useful for denoising natural image patches. 
This suggests that during the meta-learning process learns to use general features that are useful for denoising in a wide variety of settings.

\subsection{Compressive Sensing}
\begin{figure}[t]
    \centering
    \includegraphics[width=\linewidth]{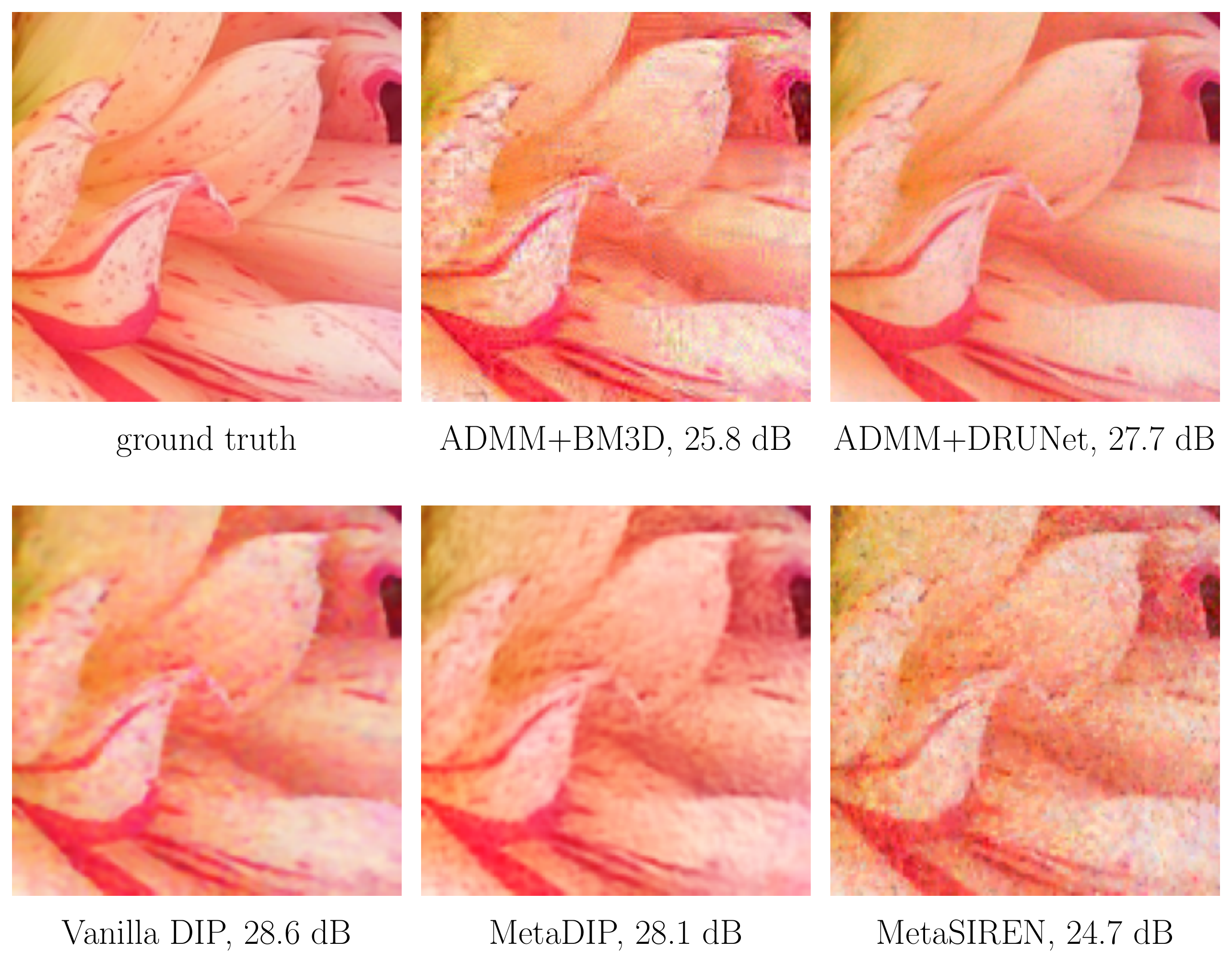}
	\vspace{-20pt}
    \caption{Various methods for solving compressed sensing problems applied to problems at $\sigma=25$ and $\frac{m}{n} = \frac{1}{4}$ involving compressed sensing using an image from the natural images dataset. }
    \label{fig:cs_nat}
\end{figure}

Compressive sensing refers to the inverse problem where the measurements are of the form 
\begin{equation}
    y = Ax + e,
\end{equation}
where $y \in \mathbb{R}^m$, $A \in \mathbb{R}^{m \times n}$, $x \in \mathbb{R}^n$, and $e \in \mathbb{R}^n$ for $m \ll n$~\cite{ongie2020deep}.
Here, we study the instance where the entries are i.i.d.~selected from $N(0, 1/m)$ and $e$ is additive white Gaussian noise with known variance $\sigma^2$.
For natural image patches, the qualitative results are presented in Figure \ref{fig:cs_nat} and the quantitative results in Table \ref{tab:cs_patch}.

\setlength{\tabcolsep}{2pt}
\begin{table}[t]
	\begin{tabular}{lrrr|rrr}
  \toprule
		\multirow{2}{*}{Method}& 
		\multicolumn{3}{c|}{$\frac{m}{n} = \frac{1}{4}$} & \multicolumn{3}{c}{$\frac{m}{n} = \frac{1}{8}$} \\
		\cmidrule{2-7} 
		& \shortstack{PSNR \\ (faces)} & \shortstack{PSNR \\ (natural)} & Time & \shortstack{PSNR \\ (faces)} & \shortstack{PSNR \\ (natural)} & Time  \\
		\midrule
		ADMM+BM3D & 26.0 & 26.8 & 35.3 & 27.1 & 25.7 & 70.5\\
		ADMM+DRUNet & 27.6 & 27.2 & 0.58 & 28.1 & 27.7 & 1.1 \\ 
		DIP (1000) & 29.4 & 29.2 & 24.9 & 28.5 & 28.3 & 23.2 \\
		DIP (50) & 9.84 & 10.0 & 1.2 & 9.83 & 10.0 & 1.2 \\
		MetaSIREN & 26.5 & 25.1 & 0.53 & 23.7 & 22.0 & 0.73 \\
		MetaDIP & 29.0 & 28.3 & 1.1 & 27.2 & 26.7 & 2.0  \\
		\bottomrule
	\end{tabular}%
	\caption{Average PSNRs (dB) and runtimes (seconds) of methods for solving compressive sensing problems involving $128 \times 128$ natural image patches for various downsampling factors.}
	\label{tab:cs_patch}%
\hfill
\end{table}%

MetaDIP is much faster than Vanilla DIP and PnP ADMM with BM3D, while being better than PnP ADMM in most settings and having comparable performance with Vanilla DIP.
It is slower than Meta+SIREN and PnP ADMM with DRUNet, but has better performance than Meta+SIREN and is occasionally better than PnP ADMM with DRUNet. 
The results imply that MetaDIP has a strong performance/runtime split across various compressive sensing settings compared to the baselines.

\subsection{Compressive Phase Retrieval}
\begin{figure}[t]
    \centering
    \includegraphics[width=\linewidth]{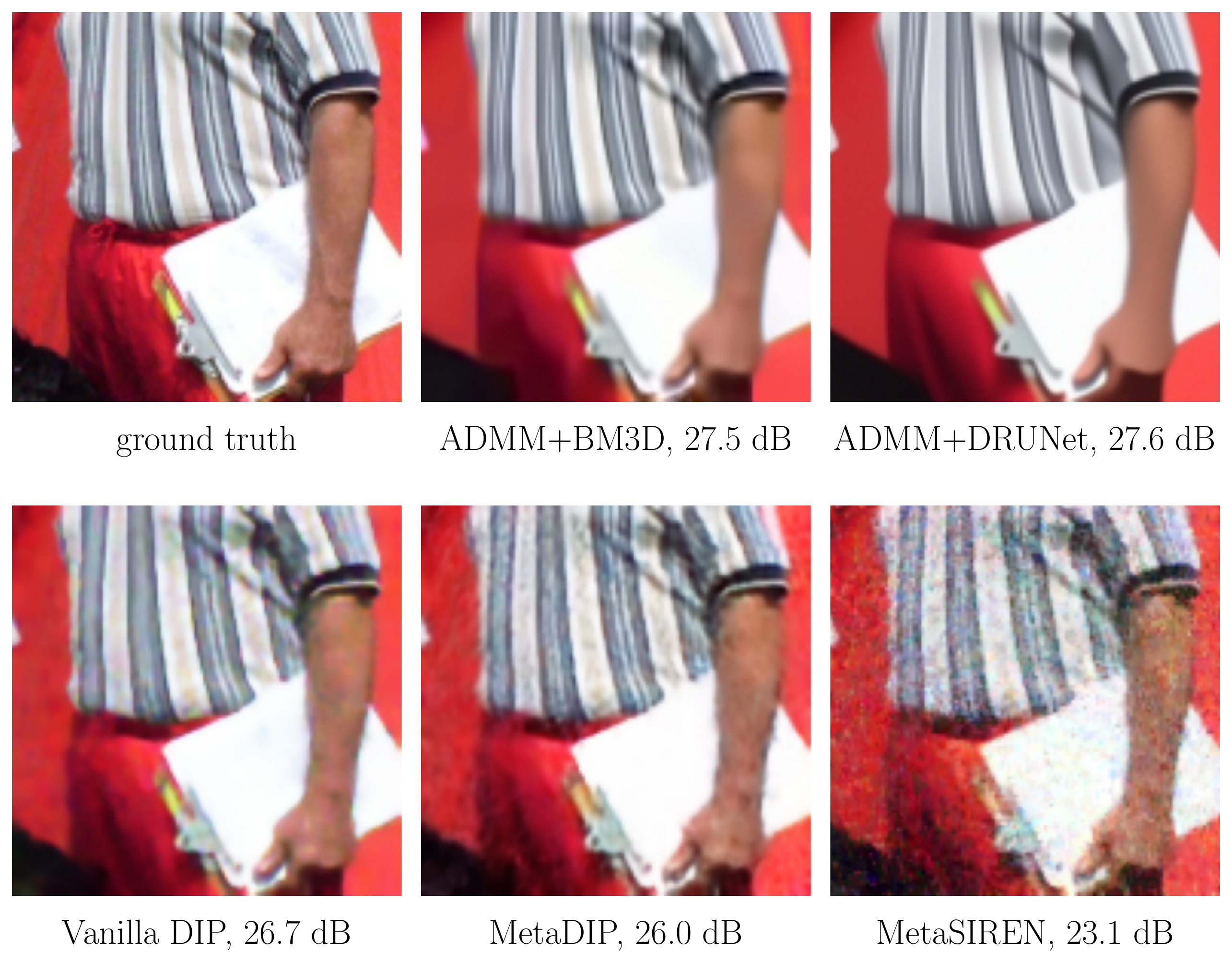}
	\vspace{-20pt}
    \caption{Various methods for solving compressive phase retrieval problems applied to problems at $\sigma=25$ and $\frac{m}{n} = \frac{1}{4}$ involving compressed sensing using an image from the natural images dataset. }
    \label{fig:cpr_nat}
\end{figure}

Compressive phase retrieval refers to the inverse problem where the measurements are of the form 
\begin{equation}
    y = |Ax + e|.
\end{equation}
Here, we study the instance where the entries are i.i.d. selected from $CN(0, 1/m)$ and $e$ is additive white Gaussian noise with known variance $\sigma$.
Note that compressive phase retrieval is a nonlinear inverse problem, which is typically more difficult and computationally expensive to solve than standard compressive sensing.  
Qualitative results are presented in Figure~\ref{fig:cpr_nat} and quantitative results are summarized in Table~\ref{tab:pr_patch}.

\begin{table}[t]
	\ninept
	\parbox{\linewidth}{
	\centering
	\begin{tabular}{lrrr|rrr}
  \toprule
		\multirow{2}{*}{Method}& 
		\multicolumn{3}{c|}{$\frac{m}{n} = \frac{1}{4}$} & \multicolumn{3}{c}{$\frac{m}{n} = \frac{1}{8}$} \\
		\cmidrule{2-7} 
		& \shortstack{PSNR \\ (faces)} & \shortstack{PSNR \\ (natural)} & Time & \shortstack{PSNR \\ (faces)} & \shortstack{PSNR \\ (natural)} & Time  \\
		\midrule
		ADMM+BM3D & 29.4 & 29.0 & 40.3 & 28.4 & 27.7 & 75.0 \\
		ADMM+DRUNet & 29.1 & 29.1 & 5.7 & 30.2 & 30.2 & 5.8 \\ 
		DIP (1000) & 29.3 & 29.3 & 38.5 & 28.4 & 28.3 & 25.2 \\
		DIP (50) & 9.9 & 9.9 & 1.9 & 10.2 & 10.1 & 1.3\\
		MetaSIREN & 27.4 & 25.7 & 1.6 & 24.7 & 22.5 & 1.8 \\
		MetaDIP & 29.8 & 28.7 & 1.8 & 28.6 & 27.6 & 2.2 \\
		\bottomrule
	\end{tabular}%
	\caption{Average PSNRs (dB) and runtimes (seconds) of methods for solving compressive phase retrieval problems involving $128 \times 128$ natural image patches for various downsampling factors.}
	\label{tab:pr_patch}%
	}
\hfill
\end{table}%
MetaDIP achieves the best balance of speed and accuracy: It runs far faster than all methods except MetaSIREN and is far more accurate than MetaSIREN.
A part of the reason for MetaDIP's favorable speed is the difficulty of solving a least squares subproblem for the ADMM updates for the compressive phase retrieval problem. 
These results illustrate the utility of MetaDIP in solving complicated nonlinear inverse problem like compressive phase retrieval. 
\vspace{-10pt}
\section{Conclusion}
\vspace{-5pt}
\label{sec:conclusion}
In this work, we propose to use meta-learning to learn a better initialization of DIP, which drastically reduces the required number of iterations and wall time for solving inverse problems in imaging.
Our experiment results shows that our method, MetaDIP, significantly outperforms MAML-initialized SIREN. While its denoising performance trails that of state-of-the-art methods, when solving compressive inverse problems MetaDIP is competitive with PnP methods, relatively fast, and far easier to tune. 
These result demonstrate that MetaDIP is an efficient, effective, and easy-to-apply method for solving inverse problems in imaging.

\bibliographystyle{IEEEbib}
\bibliography{strings,refs}

\begin{thebibliography}{10}

\bibitem{ongie2020deep}
Gregory Ongie, Ajil Jalal, Christopher~A. Metzler, Richard~G. Baraniuk,
  Alexandros~G. Dimakis, and Rebecca Willett,
\newblock ``Deep learning techniques for inverse problems in imaging,''
\newblock {\em IEEE Journal on Selected Areas in Information Theory}, vol. 1,
  no. 1, pp. 39--56, 2020.

\bibitem{8579082}
Victor Lempitsky, Andrea Vedaldi, and Dmitry Ulyanov,
\newblock ``Deep image prior,''
\newblock in {\em 2018 IEEE/CVF Conference on Computer Vision and Pattern
  Recognition}, 2018, pp. 9446--9454.

\bibitem{van2018compressed}
Dave Van~Veen, Ajil Jalal, Mahdi Soltanolkotabi, Eric Price, Sriram Vishwanath,
  and Alexandros~G Dimakis,
\newblock ``Compressed sensing with deep image prior and learned
  regularization,''
\newblock {\em arXiv preprint arXiv:1806.06438}, 2018.

\bibitem{liu2021solving}
Jiapeng Liu, Muralidhar~M Balaji, Christopher~A Metzler, M~Salman Asif, and
  Prasanna Rangarajan,
\newblock ``Solving inverse problems using self-supervised deep neural nets,''
\newblock in {\em Computational Optical Sensing and Imaging}. Optical Society
  of America, 2021, pp. CTh5A--2.

\bibitem{tancik2020meta}
Matthew Tancik, Ben Mildenhall, Terrance Wang, Divi Schmidt, Pratul~P.
  Srinivasan, Jonathan~T. Barron, and Ren Ng,
\newblock ``Learned initializations for optimizing coordinate-based neural
  representations,''
\newblock in {\em CVPR}, 2021.

\bibitem{heckel2018deep}
Reinhard Heckel and Paul Hand,
\newblock ``Deep decoder: Concise image representations from untrained
  non-convolutional networks,''
\newblock in {\em International Conference on Learning Representations}, 2018.

\bibitem{bostan2020deep}
Emrah Bostan, Reinhard Heckel, Michael Chen, Michael Kellman, and Laura Waller,
\newblock ``Deep phase decoder: self-calibrating phase microscopy with an
  untrained deep neural network,''
\newblock {\em Optica}, vol. 7, no. 6, pp. 559--562, 2020.

\bibitem{finn2017model}
Chelsea Finn, Pieter Abbeel, and Sergey Levine,
\newblock ``Model-agnostic meta-learning for fast adaptation of deep
  networks,''
\newblock in {\em International Conference on Machine Learning 2017}.

\bibitem{Sitzmann2020MetaSDFMS}
Vincent Sitzmann, Eric Chan, Richard Tucker, Noah Snavely, and Gordon
  Wetzstein,
\newblock ``Metasdf: Meta-learning signed distance functions,''
\newblock {\em ArXiv}, vol. abs/2006.09662, 2020.

\bibitem{bergman2021metanlr}
Alexander~W. Bergman, Petr Kellnhofer, and Gordon Wetzstein,
\newblock ``Fast training of neural lumigraph representations using meta
  learning,''
\newblock in {\em NeurIPS}, 2021.

\bibitem{sitzmann2019siren}
Vincent Sitzmann, Julien~N.P. Martel, Alexander~W. Bergman, David~B. Lindell,
  and Gordon Wetzstein,
\newblock ``Implicit neural representations with periodic activation
  functions,''
\newblock in {\em Proc. NeurIPS}, 2020.

\bibitem{pmlr-v97-wu19d}
Yan Wu, Mihaela Rosca, and Timothy Lillicrap,
\newblock ``Deep compressed sensing,''
\newblock in {\em Proceedings of the 36th International Conference on Machine
  Learning}. 09--15 Jun 2019, vol.~97 of {\em Proceedings of Machine Learning
  Research}, pp. 6850--6860, PMLR.

\bibitem{pmlr-v97-rahaman19a}
Nasim Rahaman, Aristide Baratin, Devansh Arpit, Felix Draxler, Min Lin, Fred
  Hamprecht, Yoshua Bengio, and Aaron Courville,
\newblock ``On the spectral bias of neural networks,''
\newblock in {\em Proceedings of the 36th International Conference on Machine
  Learning}. 09--15 Jun 2019, vol.~97 of {\em Proceedings of Machine Learning
  Research}, pp. 5301--5310, PMLR.

\bibitem{NEURIPS2020_55053683}
Matthew Tancik, Pratul Srinivasan, Ben Mildenhall, Sara Fridovich-Keil, Nithin
  Raghavan, Utkarsh Singhal, Ravi Ramamoorthi, Jonathan Barron, and Ren Ng,
\newblock ``Fourier features let networks learn high frequency functions in low
  dimensional domains,''
\newblock in {\em Advances in Neural Information Processing Systems}, 2020,
  vol.~33, pp. 7537--7547.

\bibitem{antoniou2018train}
Antreas Antoniou, Harrison Edwards, and Amos Storkey,
\newblock ``How to train your maml,''
\newblock {\em arXiv preprint arXiv:1810.09502}, 2018.

\bibitem{Rumelhart:1986we}
David~E. Rumelhart, Geoffrey~E. Hinton, and Ronald~J. Williams,
\newblock ``{Learning Representations by Back-propagating Errors},''
\newblock {\em Nature}, vol. 323, no. 6088, pp. 533--536, 1986.

\bibitem{8682856}
Jiaming Liu, Yu~Sun, Xiaojian Xu, and Ulugbek~S. Kamilov,
\newblock ``Image restoration using total variation regularized deep image
  prior,''
\newblock in {\em ICASSP 2019 - 2019 IEEE International Conference on
  Acoustics, Speech and Signal Processing (ICASSP)}, 2019, pp. 7715--7719.

\bibitem{liu2015faceattributes}
Ziwei Liu, Ping Luo, Xiaogang Wang, and Xiaoou Tang,
\newblock ``Deep learning face attributes in the wild,''
\newblock in {\em Proceedings of International Conference on Computer Vision
  (ICCV)}, December 2015.

\bibitem{MartinFTM01}
D.~Martin, C.~Fowlkes, D.~Tal, and J.~Malik,
\newblock ``A database of human segmented natural images and its application to
  evaluating segmentation algorithms and measuring ecological statistics,''
\newblock in {\em Proc. 8th Int'l Conf. Computer Vision}, July 2001, vol.~2,
  pp. 416--423.

\bibitem{ma2017waterloo}
Kede Ma, Zhengfang Duanmu, Qingbo Wu, Zhou Wang, Hongwei Yong, Hongliang Li,
  and Lei Zhang,
\newblock ``{Waterloo Exploration Database}: New challenges for image quality
  assessment models,''
\newblock {\em IEEE Transactions on Image Processing}, vol. 26, no. 2, pp.
  1004--1016, Feb. 2017.

\bibitem{Agustsson_2017_CVPR_Workshops}
Eirikur Agustsson and Radu Timofte,
\newblock ``Ntire 2017 challenge on single image super-resolution: Dataset and
  study,''
\newblock in {\em The IEEE Conference on Computer Vision and Pattern
  Recognition (CVPR) Workshops}, July 2017.

\bibitem{Lim_2017_CVPR_Workshops}
Bee Lim, Sanghyun Son, Heewon Kim, Seungjun Nah, and Kyoung~Mu Lee,
\newblock ``Enhanced deep residual networks for single image
  super-resolution,''
\newblock in {\em The IEEE Conference on Computer Vision and Pattern
  Recognition (CVPR) Workshops}, July 2017.

\bibitem{9454311}
Kai Zhang, Yawei Li, Wangmeng Zuo, Lei Zhang, Luc Van~Gool, and Radu Timofte,
\newblock ``Plug-and-play image restoration with deep denoiser prior,''
\newblock {\em IEEE Transactions on Pattern Analysis and Machine Intelligence},
  pp. 1--1, 2021.

\bibitem{DBLP:journals/corr/KingmaB14}
Diederik~P. Kingma and Jimmy Ba,
\newblock ``Adam: {A} method for stochastic optimization,''
\newblock in {\em 3rd International Conference on Learning Representations,
  {ICLR} 2015}, 2015.

\bibitem{10.1117/12.643267}
Kostadin Dabov, Alessandro Foi, Vladimir Katkovnik, and Karen Egiazarian,
\newblock ``{Image denoising with block-matching and 3D filtering},''
\newblock in {\em Image Processing: Algorithms and Systems, Neural Networks,
  and Machine Learning}. International Society for Optics and Photonics, 2006,
  vol. 6064, pp. 354 -- 365, SPIE.

\bibitem{venkatakrishnan2013plug}
Singanallur~V Venkatakrishnan, Charles~A Bouman, and Brendt Wohlberg,
\newblock ``Plug-and-play priors for model based reconstruction,''
\newblock in {\em 2013 IEEE Global Conference on Signal and Information
  Processing}. IEEE, 2013, pp. 945--948.

\bibitem{wei2020tuning}
Kaixuan Wei, Angelica Aviles-Rivero, Jingwei Liang, Ying Fu, Carola-Bibiane
  Sch{\"o}nlieb, and Hua Huang,
\newblock ``Tuning-free plug-and-play proximal algorithm for inverse imaging
  problems,''
\newblock in {\em International Conference on Machine Learning}. PMLR, 2020,
  pp. 10158--10169.

\end{thebibliography}

\end{document}